\pdfoutput=1

\documentclass[11pt]{article}

\usepackage[]{acl}

\usepackage{times}
\usepackage{latexsym}
\usepackage{color}
\usepackage{pgfplots}
\pgfplotsset{width=10cm,compat=1.16}
\usepackage{subcaption}

\usepackage{amsmath}
\usepackage{listings}

\expandafter\def\expandafter\normalsize\expandafter{%
    \normalsize%
    \setlength\abovedisplayskip{-5pt}%
    \setlength\belowdisplayskip{-9pt}%
    \setlength\abovedisplayshortskip{0pt}%
    \setlength\belowdisplayshortskip{0pt}%
}

\usepackage{mathtools}
\usepackage{algorithm,algorithmic}
\usepackage{cases}

\usepackage[T1]{fontenc}

\usepackage[utf8]{inputenc}

\usepackage{microtype}
\usepackage{url}

\usepackage{multirow}
\usepackage{tabularx}
\usepackage{lipsum} 
\usepackage{booktabs}
\usepackage{inconsolata}
\usepackage{graphicx}
\usepackage{calligra}
\usetikzlibrary{pgfplots.polar}

\definecolor{color1}{RGB}{159,168,218}
\definecolor{color2}{RGB}{173,201,198}
\definecolor{color3}{RGB}{207,216,220}
\definecolor{color4}{RGB}{247,194,183}
\definecolor{color5}{RGB}{187,230,190}
\definecolor{color6}{RGB}{187,230,222}
\definecolor{darkgreen}{rgb}{0,0.5,0}
\definecolor{fancy_color1}{RGB}{51,75,127}
\definecolor{fancy_color2}{RGB}{180,75,144}
\definecolor{pie_pink}{HTML}{feaba7}
\definecolor{pie_rose}{HTML}{fbe1ed}
\definecolor{pie_purple}{HTML}{dbd1ed}
\definecolor{pie_green}{HTML}{adeaad}
\definecolor{pie_orange}{HTML}{ffcd88}
\definecolor{pie_gray}{HTML}{dde7eb}
\definecolor{orchid}{RGB}{218, 112, 214}

\definecolor{radar_color1}{HTML}{0077ad}
\definecolor{radar_color2}{HTML}{aa71c7}
\definecolor{radar_color3}{HTML}{ff7f4f}
\definecolor{radar_color4}{HTML}{e765af}
\definecolor{radar_color5}{HTML}{6178c5}
\definecolor{radar_color6}{HTML}{ffa600}
\definecolor{radar_color7}{HTML}{ff6684}

\definecolor{bar_color1}{HTML}{0099de}
\definecolor{bar_color2}{HTML}{7a92ee}
\definecolor{bar_color3}{HTML}{c684e5}
\definecolor{bar_color4}{HTML}{ff72c2}
\definecolor{bar_color5}{HTML}{ff6e8d}
\definecolor{bar_color6}{HTML}{ff8453}
\definecolor{bar_color7}{HTML}{ffa600}

\title{Leveraging Large Language Models for \\ Concept Graph Recovery and Question Answering in NLP Education}

\author{Rui Yang$^{1*}$, Boming Yang$^2$, Sixun Ouyang$^3$, Tianwei She$^3$, Aosong Feng$^4$,\\ \textbf{Yuang Jiang$^3$, Freddy Lecue$^5$, Jinghui Lu$^3$, Irene Li$^{2,3*}$} \\
\\
$^1$Duke-NUS Medical School, $^2$University of Tokyo, $^3$Smartor Inc., \\
$^4$Yale University, $^5$INRIA \\
yang\_rui@u.nus.edu, ireneli@ds.itc.u-tokyo.ac.jp
}

\begin{document}
\maketitle
\begin{abstract}

In the domain of Natural Language Processing (NLP), Large Language Models (LLMs) have demonstrated promise in text-generation tasks. However, their educational applications, particularly for domain-specific queries, remain underexplored. This study investigates LLMs' capabilities in educational scenarios, focusing on concept graph recovery and question-answering (QA). We assess LLMs' zero-shot performance in creating domain-specific concept graphs and introduce \textbf{TutorQA}, a new expert-verified NLP-focused benchmark for scientific graph reasoning and QA. TutorQA consists of five tasks with 500 QA pairs. To tackle TutorQA queries, we present CGLLM, a pipeline integrating concept graphs with LLMs for answering diverse questions. Our results indicate that LLMs' zero-shot concept graph recovery is competitive with supervised methods, showing an average 3\% F1 score improvement. In TutorQA tasks, LLMs achieve up to 26\% F1 score enhancement. Moreover, human evaluation and analysis show that CGLLM generates answers with more fine-grained concepts.

\end{abstract}

\section{Introduction}
Recently, Large Language Models (LLMs) such as GPT~\cite{achiam2023gpt} and LLaMa~\cite{touvron2023llama} have shown superior performance in many tasks in the field of natural language processing (NLP), especially in generation tasks, including question answering (QA), text summarization, machine translation and beyond~\cite{Tan2023CanCR,Goyal2022NewsSA,Liu2023RecPromptAP}. As a result, such techniques have also been applied to the subfield of educational applications, including teaching assistants and chatbots~\cite{KhanAcademy2023,Botpress2023}.

\begin{figure}[t]
    \centering
    \includegraphics[width=0.98\linewidth]{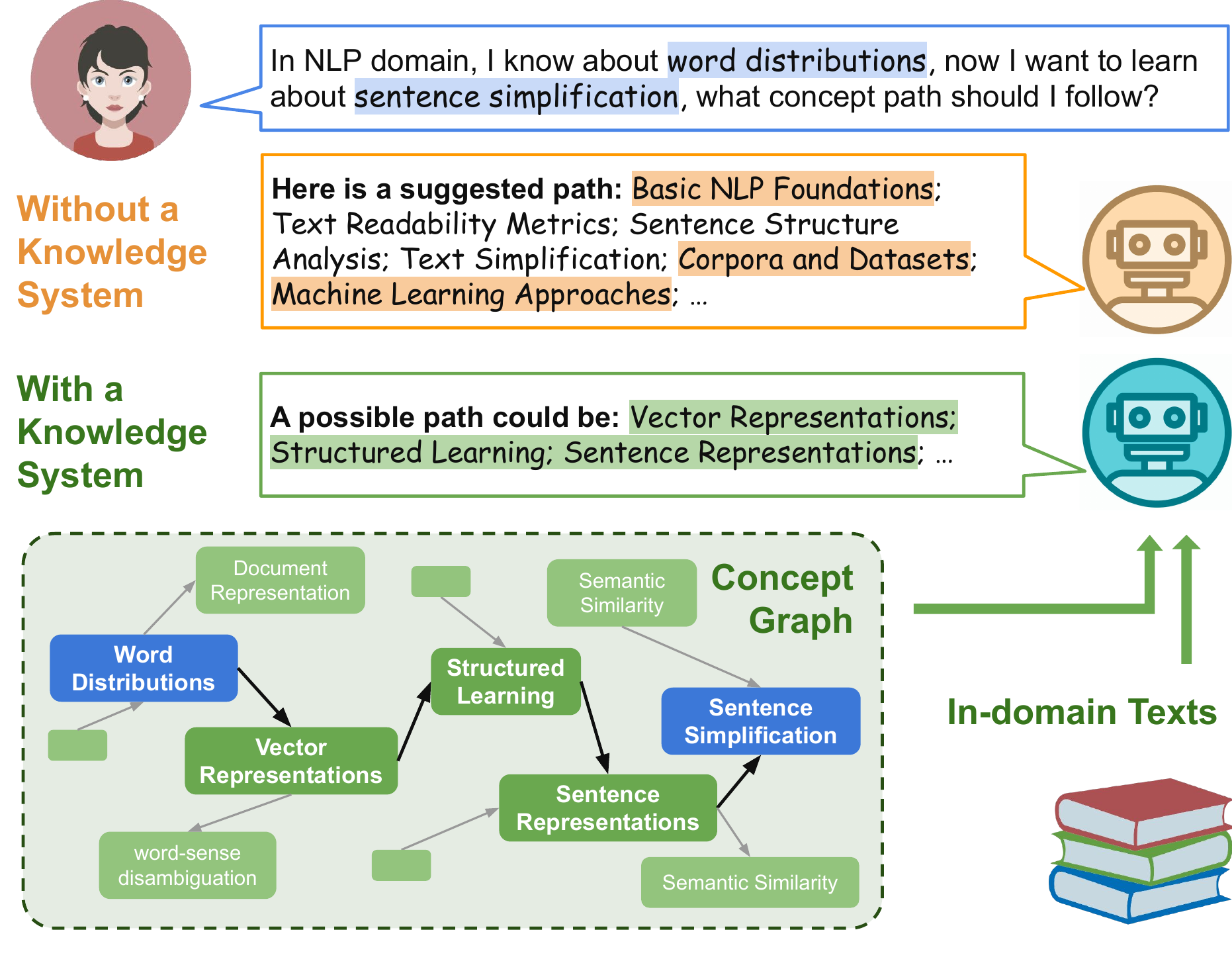}
    \caption{Comparison of QA systems with and without a Knowledge System. }
    \label{fig:intro}
    \vspace{-5mm}
\end{figure}

Studies have demonstrated that the use of virtual teaching assistants positively impacts higher education~\cite{Essel2022TheIO, okonkwo2020python}. Particularly, recent applications, such as LLM-powered chatbots, are increasingly being utilized in educational settings, including in areas like the Medical Licensing Examination~\cite{Gilson2023HowDC}, college-level scientific problems~\cite{Wang2023SciBenchEC}, college-level NLP exam questions~\cite{Song2023NLPBenchEL}, and Geographic Information Systems (GIS) exams~\cite{Mooney2023TowardsUT}. While most studies focus on the end-to-end evaluation of LLMs' question-answering capabilities and overall knowledge, the importance of a learning system's awareness, e.g., a learning map, is often overlooked.  As illustrated in Figure~\ref{fig:intro}, a user asks a question involving specific concepts (highlighted in blue). Ideally, the response should reflect the relationships between these concepts, essentially outlining the learning path that connects them. Without a knowledge system (indicated in orange), an LLM might offer answers that are somewhat relevant but too general, including broad concepts like \lq Basic NLP Foundations' and \lq Machine Learning Approach', or it may introduce confusing concepts with incorrect specificity, such as \lq Corpora and Datasets', which lack precise relevance to the field. In contrast, when equipped with a knowledge system---as demonstrated by a concept graph showing prerequisite relationships---supplemented by relevant in-domain texts, the response becomes more refined, reflecting a deeper understanding of the concept relationships informed by the knowledge system.

While LLMs are often regarded as black boxes, the generated content can occasionally include hallucinations or exhibit issues with factuality or interpretability~\cite{Wang2023FactcheckGPTEF, Manakul2023SelfCheckGPTZB, Huang2023CanLL}. 
To address these challenges, a way is to integrate knowledge graphs (KGs) into LLMs.~\citet{Yang2023ChatGPTIN} and~\citet{Yang2023IntegratingUK} have enhanced LLMs for text generation by incorporating existing knowledge graphs, thus improving factuality in text generation. Additionally, other studies have explored the augmentation of LLMs with KGs for various KG tasks, including knowledge graph completion, reasoning, and more~\cite{Zhu2023LLMsFK, Chen2023IncorporatingSS}. Our work proposes the application of utilizing an LLM-generated concept graph~\cite{Li2022VariationalGA} to answer more complex and challenging questions from an educational perspective. 

In this study, we utilize LLMs to recover concept graphs, demonstrating their ability in supporting downstream question-answering tasks in NLP education. Specifically, our contributions are summarized as follows: 

\begin{itemize}
\setlength\itemsep{-0.5em}
\item{We explore the capability of LLMs in recovering scientific concept graphs, employing various prompt strategies.} 

\item{We introduce TutorQA, a benchmark designed for scientific graph reasoning and question-answering. This benchmark comprises 500 expert-verified question-answering/reference pairs, spanning five distinct tasks.} 

\item{We develop a pipeline, CGLLM, which enhances the interaction between LLMs and the concept graph for TutorQA. Our results indicate that CGLLM significantly enhances performance, achieving up to a 26\% increase in the F1 score.\footnote{Code and data can be found in \url{https://github.com/IreneZihuiLi/CGPrompt}.}}

\end{itemize}

To the best of our knowledge, our work is pioneering in the creation of scientific concept graphs using the zero-shot capabilities of LLMs. Additionally, it sets a precedent in benchmarking concept graph reasoning and text generation within a specific domain and college-level education, verifying the efficacy of the proposed approach.


\section{Related Work}

\begin{figure*}[t!]
    \centering
\includegraphics[width=0.95\textwidth]{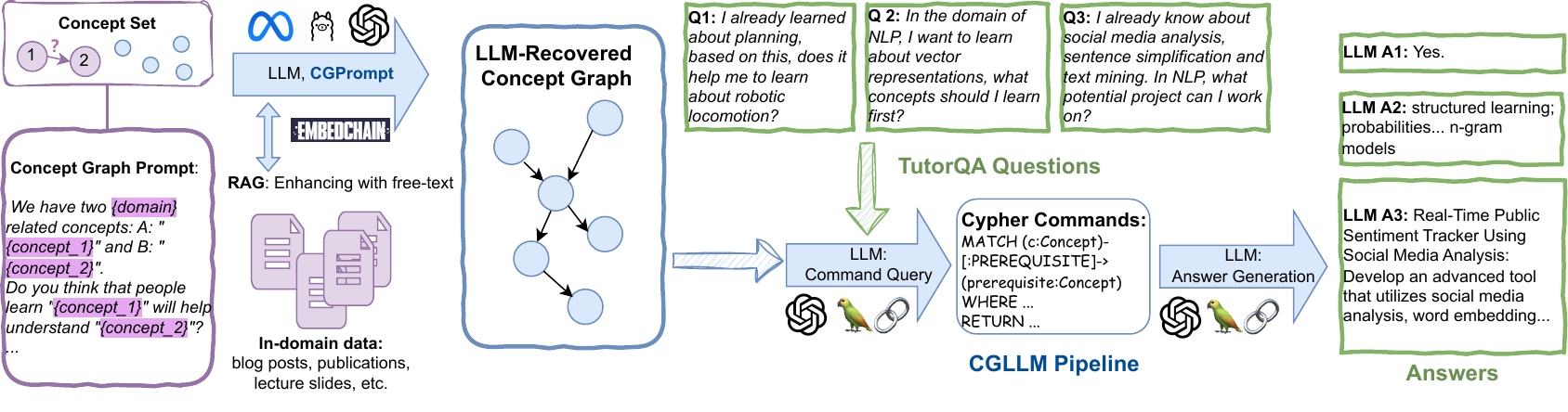}
\vspace{-2mm}
    \caption{Applying LLMs for Concept Graph Recovery and TutorQA in NLP Education: we introduce CGPrompt, CGLLM pipeline as well as the TutorQA benchmark. }
    \label{fig:model}
    \vspace{-3mm}
\end{figure*}

\paragraph{Knowledge Graph} Research on knowledge graphs (KGs) often targets general domains, such as Wikidata-based projects~\cite{wikidata}, emphasizing entity relationships. There's also significant work on Scientific Knowledge Graphs (SKGs) and citation graphs~\cite{LeTuan2022TowardsBL,Kalla2023ScientificKG,Ahrabian2023PubGraphAL}, mapping connections in academic publications. Other specialized KGs include Medical KGs~\cite{Wu2023MedicalKG,Murali2023TowardsEH} for healthcare, Legal KGs~\cite{Gao2023HowLK,Liu2022ResearchOT} for law, and Financial KGs~\cite{Tang2023IntelligentSR,Kertkeidkachorn2023FinKGAC} for finance. Each domain-specific KG is vital in structuring and analyzing data within its field.

\paragraph{Concept Graph} Concept graphs are a specialized type of Knowledge Graph~\cite{Urgo2022LearningAI,Pan2017PrerequisiteRL,Yang2015ConceptGL}. These graphs typically feature nodes representing concepts or courses, interconnected by relationships such as dependencies or prerequisites~\cite{Pan2017PrerequisiteRL}. Recent studies have explored machine learning methods for link prediction in concept nodes~\cite{Jia2021HeterogeneousGN,Li2021UnsupervisedCP}. However, these methods predominantly rely on supervised or semi-supervised learning approaches~\cite{Li2020RVGAERG}, or on information extraction from extensive free-text sources~\cite{Pan2017PrerequisiteRL,Li2021UnsupervisedCP}. The former approach often requires annotations from human experts to construct training data, which can be both time-consuming and costly. Conversely, the latter approach demands high-quality text data and typically involves manually designed information extraction or additional learning processes. In this work, we investigate a zero-shot scenario utilizing LLMs to address these challenges.

\paragraph{Educational NLP} Modern NLP and Artificial Intelligence (AI) techniques have been applied to a wide range of applications, with education being a significant area. For instance, various tools have been developed focusing on writing assistance, language study, automatic grading, and quiz generation~\cite{Zhang2023VISARAH,Lu2023ErrorAP,GonzlezCarrillo2021AutomaticGT,Seyler2015GeneratingQQ}. Moreover, in educational scenarios, providing responses to students still requires considerable effort, as the questions often demand a high degree of relevance to the study materials and strong domain knowledge. Consequently, many studies have concentrated on developing automatic question-answering models~\cite{Zylich2020ExploringAQ,Hicke2023ChaTATA}, which tackle a range of queries, from logistical to knowledge-based questions. In this work, we integrate an LLM-generated concept graph for various QA tasks in NLP education.


\section{Concept Graph Recovery}
In this section, we introduce our proposed method, CGPrompt (\textbf{C}oncept \textbf{G}raph \textbf{P}rompt), that recovers a concept graph using Large Language Models. 

\subsection{Problem Definition}

For a specific domain, a concept graph is defined as $\calligra{G} = (\calligra{C}, \calligra{E})$. This graph is a directed one where the nodes, denoted by $\calligra{C}$, represent concepts, and the edges, $\calligra{E}$, signify dependency or prerequisite relations between each positive concept pair $(A, B)$. For instance, in the NLP domain, the relation "Viterbi Algorithm" $\rightarrow$ "POS Tagging" implies that to learn the concept of "POS Tagging," one must first understand "Viterbi Algorithm." Initially, a predefined set of concepts $\calligra{C}$ is given. In a zero-shot setting, the task involves inferring $\calligra{E}$, which includes all potential edges between concept pairs, in order to construct the concept graph $\calligra{G}$. In the supervised setting, some positive edges are given (from the training set), and the task is to infer the unseen edges.

\begin{table*}[t]
\centering
\begin{tabular}{lcccccccc}
\toprule
\textbf{Method} & \multicolumn{2}{c}{\textbf{NLP}} & \multicolumn{2}{c}{\textbf{CV}} & \multicolumn{2}{c}{\textbf{BIO}} & \multicolumn{2}{c}{\textbf{Overall}} \\
\cmidrule(lr){2-3} \cmidrule(lr){4-5} \cmidrule(lr){6-7} \cmidrule(lr){8-9}
 & \textbf{Acc} & \textbf{F1} & \textbf{Acc} & \textbf{F1} & \textbf{Acc} & \textbf{F1} & \textbf{Acc} & \textbf{F1} \\
\midrule 
\multicolumn{9}{l}{\textit{Supervised Baselines}} \\
P2V & 0.6369 & 0.5961 & 0.7642 & 0.7570 & 0.7200 & 0.7367 & 0.7070 & 0.6966 \\
BERT & 0.7088 & 0.6963 & 0.7572 & 0.7495 & 0.7067 & 0.7189 & 0.7242 & 0.7216 \\
DeepWalk & 0.6292 & 0.5860 & 0.7988 & 0.7910 & 0.7911 & 0.8079 & 0.7397 & 0.7283 \\
Node2vec & 0.6209 & 0.6181 & 0.8197 & 0.8172 & 0.7956 & 0.8060 & 0.7454 & 0.7471 \\
\midrule[0.2pt]
\multicolumn{9}{l}{\textit{LLM - Supervised}} \\
LLaMa - Classifier & 0.6684 & 0.6475 & 0.6184 & 0.6549 & 0.6043 & 0.6644 & 0.6304 & 0.6556 \\
LLaMa - Graph & \underline{0.7174} & \underline{0.7673} & \underline{0.7839} & \underline{0.8223} & \underline{0.8217} & \underline{0.8464} & \underline{0.7743} & \underline{0.8120}\\
\midrule
\multicolumn{9}{l}{\textit{Zero-shot (zs)}} \\
LLaMa & 0.6058 & 0.6937 & 0.6092 & 0.6989 & 0.6261 & 0.6957 & 0.6137 & 0.6961 \\
GPT3.5 & 0.6123 & 0.7139 & 0.6667 & 0.7271 & 0.6696 & 0.6801 & 0.6495 & 0.7070 \\
GPT4 & 0.7639 & 0.7946 & \textbf{0.7391 }& \textbf{0.7629} & \textbf{0.7348} & \textbf{0.7737} & \textbf{0.7459} & \textbf{0.7771} \\
\midrule                            
\multicolumn{9}{l}{\textit{Zero-shot + RAG}} \\
GPT3.5 & 0.7587 & 0.7793 & 0.6828 & 0.7123 & 0.6870 & 0.7006 & 0.7095 & 0.7307 \\
GPT4 & \textbf{0.7755} & \textbf{0.7958} & 0.7230 & 0.7441 & 0.7174 & 0.7200 & 0.7386 & 0.7533 \\
\bottomrule
\end{tabular}
\caption{Main results across all domains and overall performance on the test set: We present accuracy (Acc) and F1 scores. Bolded figures indicate the best performance in the zero-shot setting, while underlined scores represent the highest achievements in the supervised setting. In this paper, we apply LLaMa2-70b for all experiments.}
\label{tab:main}
\vspace{-3mm}
\end{table*}

\subsection{Methodology}
We illustrate the method in Fig.~\ref{fig:model}. To recover the concept graph in a zero-shot setting, we leverage several prompt strategies, such as zero-shot, Chain-of-Thought (CoT)~\cite{CoT}, and Retriever Augmented Generation (RAG)~\cite{RAG}. Additionally, we compare these strategies with supervised methods.

\paragraph{CG Prompt} The core part of the prompt is to provide the domain name, the definition and description of the dependency relation to be predicted, and the query concepts:

\begin{lstlisting}[language={}, captionpos=b, label={lst: zs}]
 We have two {domain} related concepts: A: {concept_1} and B: {concept_2}.
Do you think that people learning {concept_1} will help in understanding {concept_2}?
Hints:
1. Answer YES or NO only.
2. This is a directional relation, which means if YES, (B,A) may be False, but (A,B) is True.
3. Your answer will be used to create a knowledge graph.
{Additional Information}
\end{lstlisting}

We also explore whether additional information, such as concept definitions from Wikipedia and neighboring concepts from training data (when available), would be beneficial. More examples including CoT prompts are provided in Appendix~\ref{app:prompt}.


\paragraph{Retriever Augmented Generation (RAG)} RAG models have shown to improve on existing LLMs on text generation tasks such as question answering~\cite{Krishna2023PromptGT}, so we further apply RAG to enhance the current pipeline. We apply a large corpus, TutorialBank~\cite{fabbri-etal-2018-tutorialbank}, to be the external data source. It contains manually collected and categorized resources for NLP, AI, and related fields. We include data details in Appendix~\ref{app:tutorialbank}.

\paragraph{Supervised Methods with LLMs} In this setting, we are given some training labels, and we conduct binary classification on unseen concept pairs. Specifically, we first obtain LLM embedding for each concept using the following prompt: \texttt{What's the definition of \{concept\}?}

We propose two models: binary classifiers and graph convolutional networks (GCN)~\cite{Kipf2016SemiSupervisedCW}.  For binary classifiers, we concatenate LLM embeddings of the given concept pair, and apply simple classifiers such as Na\"ive Bayesian classifier, linear regression, and random forest. In the GCN model, let $\mathbf{X}$ represent the feature matrix, where each row corresponds to a concept embedding. Let $\mathbf{A}$ be the adjacency matrix, comprising known prerequisite relations from the training set, and let $\mathbf{W}$ denote the learnable weight matrix. The graph convolution operation in the GCN can be defined as follows: $\mathbf{H}^{(l+1)} = \sigma \left( \mathbf{D}^{-\frac{1}{2}}\mathbf{AD}^{-\frac{1}{2}}\mathbf{H}^{(l)}\mathbf{W}^{(l)} \right)$, where $\mathbf{H}^{(l)}$ is the hidden state matrix at layer $l$, $\sigma$ is the activation function (e.g., ReLU), $\mathbf{D}$ is the diagonal degree matrix with $\mathbf{D}_{ii} = \sum_j \mathbf{A}_{ij}$, and $\mathbf{W}^{(l)}$ is the weight matrix at layer $l$. Following the work of~\citet{Li2021UnsupervisedCP}, we recover the rest relations using a scoring function:
$ \mathbf{\hat{A}}= \mathbf{\hat{X}}\mathbf{R}\mathbf{\hat{X}}$, where $\mathbf{R}$ is a trainable matrix, and $\mathbf{\hat{X}}$ is the last layer node representation. 



\subsection{Concept Graph Evaluation}
\label{sec:cge}
We conduct experiments using the LectureBankCD dataset~\cite{Li2021UnsupervisedCP}, which includes three domains: NLP, Computer Vision (CV), and Bioinformatics (BIO). Each domain features a distinct set of concepts (322 for NLP, 201 for CV, and 100 for BIO) along with prerequisite annotations for each concept pair. Moreover, LectureBankCD also contains texts from lecture notes for each domain. 

We benchmark against the following \textbf{Supervised Baselines}: P2V~\cite{Wu2020Phrase2VecPE}, BERT~\cite{Devlin2019BERTPO}, DeepWalk~\cite{Perozzi2014DeepWalkOL}, and Node2vec~\cite{Grover2016node2vecSF}. These methods utilize pretrained or graph-based models to encode concept embeddings and then perform binary classification to determine the presence of positive or negative edges in given concept pairs. In our LLM-based experiments, we implement three settings: \textbf{LLM-Supervised}, which involves using LLaMa embeddings for concept encoding and concatenation for binary classification; \textbf{Zero-shot (zs)}, which employs GCPrompt; and \textbf{Zero-shot + RAG}, which enhances the zs with the addition of RAG. The results on the official test sets are presented in Tab.~\ref{tab:main}. 
Overall, LLaMa-Graph outperforms all other models. In the zero-shot experiments, GPT4-zs surpasses other LLMs; however, the addition of RAG does not consistently yield improvements. This could be due to recent GPT models already having exposure to the data we used. Notably, GPT4-zs's overall performance even exceeds that of the best traditional supervised baseline, Node2vec. This suggests that LLMs are capable of recovering a domain-specific concept graph without relying on expert annotations. Consequently, we utilized the GPT4-zs model to conduct concept graph recovery on the \textit{full} NLP concept set. More details can be found in Appendix~\ref{app:cgrecovery}. 

\begin{table*}[t]
\centering
\small
\begin{tabular}{lllllll}
\toprule
\textbf{Dataset} & \textbf{Domain} & \textbf{Answer Type} & \textbf{With KG} & \textbf{Collection}\\ \midrule
CBT~\cite{Hill2015TheGP}          &   Open   &    Multiple Choice     &   No   &                Automated     \\
LectureBankCD~\cite{Li2021UnsupervisedCP}  &  NLP,CV,BIO     &   Binary      &    Yes     &     Expert-verified  & \\
FairytaleQA~\cite{Xu2022FantasticQA}          &   Open    &    Open-ended     &   No   &              Expert-verified      \\
ChaTa~\cite{Hicke2023ChaTATA}   & CS      & Free Text   & No      & Students    \\ 
ExpertQA~\cite{Malaviya2023ExpertQAEQ} & Science & Free Text & No &  Expert-verified  \\
\midrule
TutorQA (this work)  &  NLP   &   Open-ended, Entity List, Binary      &    Yes     &     Expert-verified    
\\ \bottomrule
\end{tabular}
\vspace{-2mm}
\caption{Comparison with other similar benchmarks: Educational or General Question Answering benchmarks. }
\label{tab:tabqa}
\vspace{-2mm}
\end{table*}




\section{TutorQA}
After the concept graph is recovered, we aim to assess its usefulness and quality. To this end, we introduce the TutorQA benchmark, which is a question-answering (QA) dataset designed for concept graph reasoning and text generation in the NLP domain. TutorQA comprises five categories, with each containing 100 QA pairs that have been verified by human experts. These questions go beyond simple syllabus inquiries, encompassing more extensive and challenging topics that require interaction with the completed graph, as well as proficiency in text comprehension and question answering. We list some similar benchmarks in Tab.~\ref{tab:tabqa}. While numerous open-domain question-answering benchmarks exist, our focus has been primarily on those within the scientific domain or tailored for college-level education, aligning with our objective to compare with benchmarks that can emulate a learning scenario. Among those, TutorQA is distinguished by its diversity in answer types. Moreover, it is specifically designed for college-level education and features expert-verified questions, ensuring a high standard of quality and relevance. 

\subsection{QA Tasks}

We summarize the tasks and provide example data in Tab.~\ref{tab:tutorQA}. More data statistics and information can be found in Appendix~\ref{app:tutorqa_info}. 

\begin{table*}[h]
\centering
\small
\begin{tabularx}{\textwidth}{lXp{5cm}c}
\toprule
\textbf{Task} & \textbf{Example Question} & \textbf{Example Answer/Reference} & \textbf{Evaluation} \\
\midrule
Task 1 & In the domain of natural language processing, I already learned about planning, based on this, would it be helpful for me to learn about robotic locomotion? & Yes & Accuracy, F1 \\
\midrule
Task 2 & In the domain of natural language processing, I want to learn about vector representations, what concepts should I learn first?	
 & structured learning;probabilities;natural language processing intro;word distributions;linear algebra;n-gram models
 &  Similarity based F1\\
\midrule
Task 3 & In the domain of natural language processing, I know about word distributions, now I want to learn about sentence simplification, what concept path should I follow?	
 & vector representations;structured learning;sentence representations & Similarity based F1 \\
\midrule
Task 4 & AI-Powered Legal Document Analyzer: A tool designed for legal professionals to rapidly analyze legal documents, extract relevant information, identify precedent cases, and suggest possible interpretations or legal strategies.
 & knowledge representation;information extraction;topic modeling;text similarity;inference;dependency parsing;semantic similarity & Similarity based F1 \\
\midrule
Task 5 & I already know about social media analysis, sentence simplification, word embedding, relation extraction, and text mining. In the domain of natural language processing, what potential project can I work on? Give me a possible idea. Show me the title and project description.
 & (*open answer) Real-Time Public Sentiment Tracker Using Social Media Analysis: Develop an advanced tool that utilizes social media analysis, word embedding, and relation extraction techniques for real-time public sentiment tracking... & Human Evaluation \\
\bottomrule
\end{tabularx}
\caption{TutorQA: example data and evaluation metrics of all 5 tasks.}
\label{tab:tutorQA}
\vspace{-2mm}
\end{table*}

\paragraph{Task 1: Prerequisite Prediction} This task is to answer if studying concept A would help the student to study concept B. In practice, this is a straightforward but very helpful way of improving study efficiency. We include 50 positive and 50 negative concept pairs. 

\paragraph{Task 2: Path Searching} 
This task involves path searching, where one inquires all potential paths that contains the query concept. In a typical learning scenario, students often encounter situations where they need to identify the prerequisites when encountering a new concept. The dataset includes 100 query concepts and the corresponding paths.

\paragraph{Task 3: Shortest Path Searching} This task is a natural extension of Task 2, involving the determination of a learning path from a source query concept to a target query concept. In a learning context, it often arises when a student possesses knowledge of certain concepts and seeks to explore a new one. In such cases, the student may wonder about the most efficient learning path to achieve their educational goal. In the dataset, there are 100 natural questions with concept pairs, and the corresponding concept path.

\paragraph{Task 4: Concept Advising} This task involves providing recommendations for relevant concepts based on a given project description. In an educational context, when a student embarks on a specific project, their initial challenge is to identify the necessary knowledge and techniques required for the project's successful completion. Unlike a straightforward named entity recognition task, this challenge arises because the relevant concepts may not be explicitly mentioned in the project description. For example, \textit{topic modeling} may be recommended in a news clustering and recommendation project. We include 100 diverse project proposals and the corresponding concept lists which are verified by domain experts. 

\paragraph{Task 5: Idea Hamster} This task is a reverse of Task 4, where instead of starting with project proposals and generating queries, it begins with a list of concepts, and the objective is to generate project proposals. In essence, this task revolves around the idea that once a student has acquired certain skills and knowledge, they can explore what intriguing projects they could undertake using this acquired expertise. Since this is an open-ended task, we provide questions and some model predictions.


\begin{table}[h]
\centering
\small
\begin{tabular}{@{}lcccc@{}}
\toprule
\textbf{ Model}     & \textbf{T1} &\textbf{ T2 }& \textbf{T3} & \textbf{T4} \\
       & Acc/F1 & S-F1 & S-F1 & S-F1 \\
\midrule
GPT3.5          & 0.60/0.70 & 0.67 & 0.78 & 0.59 \\
GPT3.5-CGLLM    & 0.92/0.92 & 0.70 & 0.83 & 0.71 \\
GPT4            & 0.52/0.68 & 0.80 & 0.89 & 0.69 \\
GPT4-CGLLM      & \textbf{0.94/0.94} & \textbf{0.89} & \textbf{0.85} & \textbf{0.72} \\
\bottomrule
\end{tabular}
\caption{Evaluation of TutorQA on Tasks 1, 2, 3, and 4, showcasing a comparison between GPT3.5, GPT4, and the CGLLM framework alongside the two base LLMs.}
\label{tab:tutorqa_res}
\vspace{-3mm}
\end{table}

\subsection{Methodology}
Shown in Fig.~\ref{fig:model}, we introduce CGLLM (\textbf{C}oncept \textbf{G}raph \textbf{LLM}s for Question Answering), a model that enhances the interaction between LLMs and the concept graph via Langchain framework.\footnote{\url{https://www.langchain.com/}} This model utilizes separate LLMs for command querying and answer generation. In the command query stage, an LLM independently generates Cypher commands to query the concept graph upon receiving a question, thereby retrieving concept paths. During the answer generation stage, these paths serve as contextual prompts for another LLM, enabling it to perform concept graph reasoning and generate answers.

\subsection{Evaluation}

\paragraph{Accuracy, F1} We report accuracy and F1 score for Task 1, as it is a binary classification task. 

\paragraph{Similarity-based F1} For Tasks 2, 3, and 4, the answers or references consist of a list of concepts. Generally, LLMs demonstrate creativity by answering with novel concepts, which are often composed of more contemporary and fresh words, even though they might not exactly match the words in the concept graph. Consequently, conventional evaluation metrics like keyword matching are unsuitable for these tasks. To address this, we propose the \textbf{similarity-based F1 (S-F1) score}. This metric considers semantic aspects when evaluating the predicted concepts and incorporates both precision and recall for a fairer assessment of these novel concepts. Based on the F1 score definition from Information Retrieval~\cite{cdmanningpraghavan2008introduction}, as shown in Eq.~\ref{eq:f1}. Here, $C_{ret}$ denotes the predicted concepts, while $C_{rel}$ is the relevant concepts in the concepts list.

\begin{equation}
\begin{aligned}
& Pre = \frac{C_{ret} \cap C_{rel}}{C_{ret}}, Rec = \frac{C_{ret} \cap C_{rel}}{C_{rel}} \\
\label{eq:f1}
\end{aligned}
\end{equation}



The key aspect is to ascertain how many concepts in $C_{ret}$ are considered to be matched to those in the ground truth concept list $C_{rel}$. Specifically, as shown in Eq.~\ref{eq:retrieve}, for a concept $m$ from the predicted list, and concept $n$ from the ground truth list, if the cosine similarity is larger than a threshold $\mu$, then we consider it is a match. \footnote{We set an empirical value $\mu$ to be 0.6. $E$ is the concept embedding, we conduct BERT~\cite{Devlin2019BERTPO}.}

\begin{equation}
\begin{aligned}
& C_{ret}^{m} \cap C_{rel}^{n} = \begin{cases}
        1 &\text{$cosine(E_m,E_n) > \mu$} \\
        0 &\text{otherwise}.
        \end{cases} \\
\label{eq:retrieve}
\end{aligned}
\end{equation}





We compare results for Tasks 1 to 4 in Tab.~\ref{tab:tutorqa_res}, using GPT3.5 and GPT4 as base models against our CGLLM framework. The results show consistent performance enhancement with the integration of our concept graph, especially in Task 1. This supports our hypothesis that a knowledge-enhanced system possesses superior QA capabilities and can effectively handle a variety of situations.

\paragraph{Expert evaluation} For Task 5, which involves generating open-ended answers, no reference or gold answers are provided. Nevertheless, we aim to compare the results between baselines and our model. Therefore, we opt for expert evaluation. Although there exist LLM-based evaluation metrics such as G-Eval~\cite{Liu2023GEvalNE}, we posit that the evaluation criteria for this task may differ and present unique challenges. Our primary focus centers on assessing the persuasiveness and scientific accuracy of the provided scientific details. For instance, we evaluate whether the project proposal appears feasible and enduring based on the query concepts.Our evaluation spans four perspectives, each rated on a scale of 1 to 5:
\textit{Concept Relevancy}: how well the project description aligns with the provided query concepts; \textit{Concept Coverage}: how comprehensively the project description covers the query concepts; \textit{Project Convincity}: the overall persuasiveness and viability of the project described; \textit{Scientific Factuality}: the accuracy and correctness of the scientific details presented in the project description. We asked two NLP domain experts to conduct the evaluation. The inter-rater reliability, measured by a Cohen's Kappa score, is 0.7626, indicating substantial agreement between the two experts. We compared the results from LLaMa, GPT4, and our model. The human evaluation reveals that LLaMa consistently generates irrelevant project proposals, resulting in poor coverage and relevancy scores. Both GPT4 and our model show strong performance across all dimensions, with our model having a slight advantage.

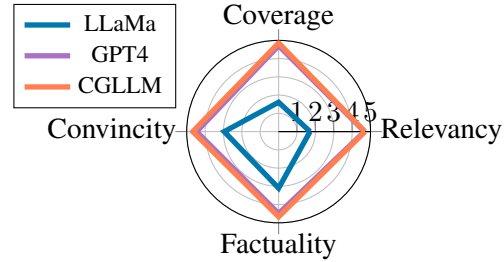
\begin{figure}
\centering
\begin{tikzpicture}
\begin{polaraxis}[
    legend style={at={(-0.5,1.2)},
        anchor=north,legend columns=1},
    legend style={font=\small}, 
    ymin=0, ymax=5,
    xtick=data,
    xticklabels={Relevancy, Coverage, Convincity, Factuality},
    ytick={1,2,3,4,5},
    width = 4cm,
    height= 4cm,
]

\addplot+[mark=none,sharp plot,line width=2pt,color=radar_color1] coordinates {
    (0,1.71)
    (90,1.61)
    (180,2.96)
    (270,3.10)
    (360,1.71)
};
\addlegendentry{LLaMa}

\addplot+[mark=none,sharp plot,line width=2pt,color=radar_color2,] coordinates {
    (0,4.67)
    (90,4.71)
    (180,4.46)
    (270,4.49)
    (360,4.67)
};
\addlegendentry{GPT4}
\addplot+[mark=none,sharp plot,line width=2pt,color=radar_color3] coordinates {
    (0,4.70)
    (90,4.86)
    (180,4.67)
    (270,4.67)
    (360,4.70)
};
\addlegendentry{CGLLM}
\end{polaraxis}
\end{tikzpicture}
\vspace{-2mm}
\caption{Human evaluation on Task 5 generations.}
\label{fig:human_eval}
\end{figure}

\section{Ablation Study on CG Recovery}

\begin{table}[t]
\centering
\small
\begin{tabular}{@{}lcccc@{}}
\toprule
  Model          & \multicolumn{2}{c}{NLP}  &  \multicolumn{2}{c}{CV} \\ 
\cmidrule(r){2-3} \cmidrule(lr){4-5} 
     & Acc & F1  & Acc & F1  \\
\midrule
GPT4 zs     & 0.7639 & 0.7946 & 0.7391&	0.7629 \\
GPT4 CoT  & 0.7342 & 0.6537 &  0.6122 &0.4159 \\
\bottomrule
\end{tabular}
\caption{Comparison of zero-shot and CoT prompts with GPT4: results on NLP and CV. }
\label{tab: cot}
\vspace{-4mm}
\end{table}

\begin{table*}[th]
\centering
\small
\begin{tabularx}{\textwidth}{lXX}
\toprule
\textbf{\textit{Question}} & \textit{I already know about \textcolor{bar_color1}{neural question answering, social media analysis, topic modeling, relation extraction, and word embedding variations}. In the domain of natural language processing, what potential project can I work on? Give me a possible idea. Show me the title and project description.} \\
\textbf{LLaMa} & \textbf{Title: }Exploring the Impact of Emojis on Textual Emotions; \textbf{Description:} Emojis have become an integral part of online communication...The project will involve the following tasks: \textcolor{bar_color6}{ 1. Data Collection: We will collect a dataset of text messages...and their corresponding emotional labels (positive, negative, neutral). 2. Data Preprocessing: We will preprocess the data by removing stop words... }\\
\textbf{GPT4} & \textbf{Title:}
Integrated Social Media Intelligence System for Trend Analysis and Relation Extraction;
\textbf{Description:}
This project aims to create a comprehensive NLP system that combines \textcolor{bar_color1}{neural question answering, social media analysis, topic modeling, relation extraction, and word embedding variations}... 
\textcolor{bar_color2}{Using neural question answering}, it will allow users to query specific trends or topics and retrieve detailed, contextually relevant insights. \textcolor{bar_color2}{Relation extraction will be used to} understand and visualize connections between different entities and themes... \\
\textbf{CGLLM} & \textbf{Title: }Multilingual Social Media Insight Engine; \textbf{Description: }The project aims to develop a comprehensive insight engine that leverages the power of \textcolor{bar_color1}{multilingual word embedding, neural question answering, and social media analysis} to provide real-time insights from social media content across different languages. By applying \textcolor{bar_color2}{topic modeling}, the engine will categorize content into various themes, making it easier for businesses and researchers to track trends and public sentiment. \textcolor{bar_color2}{Relation extraction} will be used to identify and analyze the relationships between different entities mentioned in the posts...while \textcolor{bar_color3}{document ranking} will prioritize the most relevant information for the user. \textcolor{bar_color3}{Query expansion} will improve the engine's ability to understand and respond to user queries... \\
\bottomrule
\end{tabularx}
\caption{Case study on TutorQA Task 5: LLaMa, GPT4, and GPT4-based CGLLM results (part). We highlight the \textcolor{bar_color1}{relevant concepts}, \textcolor{bar_color6}{irrelevant concepts},  \textcolor{bar_color2}{concept details} and \textcolor{bar_color3}{expanded relevant concepts}.}
\label{tab:case_study}
\vspace{-3mm}
\end{table*}


\textbf{Prompting Strategies} In Tab. ~\ref{tab: cot}, we explore the impact of different prompting strategies for concept graph recovery, comparing CoT and zero-shot prompts across both NLP and CV domains
The results indicate the introduction of CoT is not improving. We further find that CoT Prompting more frequently results in negative predictions. This finding serves as a drawback for our study, as it somewhat suppresses the performance of our system. This observation highlights the need to balance the impact of CoT on the rigor and complexity of predictions, especially in the context of graph recovery.

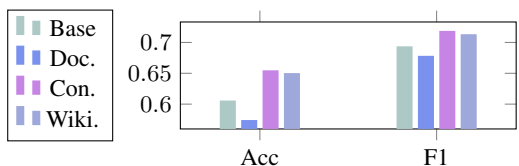
\begin{figure}[t]
    \centering
    
\begin{tikzpicture}
\centering
\small
\begin{axis}[
    ybar,
    bar width=6pt,
    enlarge x limits=0.45,
    legend style={
        at={(-0.2,0.5)}, 
        anchor=east,     
        legend columns=1 
    },
    symbolic x coords={Acc, F1},
    xtick=data,
    height=3cm,
    width=6cm,
]

\addplot[fill=color2, draw opacity=0, draw=none] coordinates {(Acc,0.6058) (F1,0.6937)};
\addlegendentry{Base}

\addplot[fill=bar_color2, draw=none, draw opacity=0] coordinates {(Acc,0.5742) (F1,0.6783) };
\addlegendentry{Doc.}

\addplot[fill=bar_color3, draw=none, draw opacity=0] coordinates {(Acc,0.6548) (F1,0.7187) };
\addlegendentry{Con.}

\addplot[fill=color1, draw=none, draw opacity=0] coordinates {(Acc,0.6503) (F1,0.7133) };
\addlegendentry{Wiki.}
\end{axis}
\end{tikzpicture}
\vspace{-3mm}
\caption{Comparison of models with external data.}
\label{fig:info_comp}
\end{figure}


\textbf{External Data} We further explore the potential of external data in enhancing concept graph recovery. This is achieved by expanding the \texttt{\{Additional Information\}} part in the \textbf{CGPrompt}. We utilize LLaMa as the \textbf{Base} model, focusing on the NLP domain. We introduce three distinct settings:

\textbf{Doc.}: Appending relevant documents from LectureBankCD, if the query concept is mentioned. 

\textbf{Con.}: Adding one-hop neighboring concepts from the training set as additional information related to the query concepts.

\textbf{Wiki.}: Incorporating the introductory paragraph of the Wikipedia page of each query concept.

As illustrated in Fig.\ref{fig:info_comp}, our findings indicate that incorporating LectureBankCD documents (Doc.) significantly diminishes performance. This decline can be attributed to the introduction of noise and excessively lengthy content, which proves challenging for the LLM to process effectively. Conversely, the inclusion of neighboring concepts (Con.) markedly enhances the base model's performance. However, it relies on training data, rendering it incompatible with our primary focus on the zero-shot setting. Incorporating Wikipedia content (Wiki.), on the other hand, also yields improvements and outperforms the use of LectureBankCD, likely due to higher text quality. 
Additionally, finetuning with external data is a viable approach, the results of which are detailed in Appendix\ref{app:ft}.


\section{TutorQA Analysis}

\begin{figure}[t]
    \centering
    \begin{tikzpicture}
    \centering
    \small
    \begin{axis}[
        ybar,
        bar width=15pt, 
        enlarge x limits=0.5, 
        legend style={
            at={(0.03,0.97)}, 
            anchor=north west 
        },
        ylabel={Count},
        ymax = 25,
        symbolic x coords={GPT3.5,GPT4},
        xtick=data,
        height=3.5cm, 
        width = 7cm,
        nodes near coords, 
        nodes near coords align={vertical},
    ]
    \addplot[fill=color2, draw=none, draw opacity=0] coordinates {(GPT3.5,3.42) (GPT4,5.24)};
    \addlegendentry{Base}
    \addplot[fill=color1, draw=none, draw opacity=0] coordinates {(GPT3.5,4.96) (GPT4,19.92)};
    \addlegendentry{CGLLM}
    \end{axis}
    \end{tikzpicture}
    \caption{Comparison of concept counts in Task 4.}
    \label{fig:task4}
\vspace{-4mm}
\end{figure}
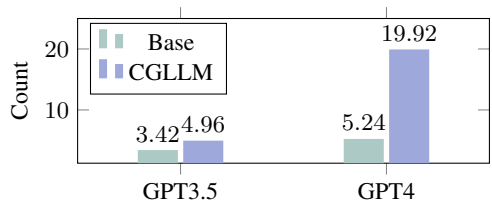

\textbf{Task 4: Concept Counts}
Task 4 involves suggesting relevant concepts based on a project description. As depicted in Fig~\ref{fig:task4}, we assess the average number of NLP concepts in responses from GPT3.5 and GPT4, along with their respective CGLLM versions. The introduction of CGLLM leads to a consistent increase in concept identification, particularly notable in GPT4, which improves from 5.24 to 19.92 concepts. Notably, the integration of the concept graph enables models to not only more accurately and effectively identify concepts related to the given topic but also to explore the prerequisites of these concepts.

\textbf{Task5: Case Study}
To further compare the outputs of the three models, we present a case study in Tab.~\ref{tab:case_study}. In this instance, the posed question incorporates five concepts (highlighted in blue), with the task being to formulate a feasible project proposal. Although LLaMa offers a substantial project description, its content and relevance to the highlighted concepts (marked in orange) are somewhat lacking. In contrast, GPT4 not only references the queried concepts but also provides detailed insights (highlighted in purple) on their potential utility within the project, such as the role of \textit{neural question answering}. Lastly, our model proposes a more comprehensive solution, elaborating on the concepts and introducing additional ones (highlighted in lavender) that come from the recovered concept graph, like \textit{document ranking} and \textit{query expansion}, while initially addressing the queried concepts. Further details about TutorQA are provided in Appendix~\ref{app:tutorqa_info}.

\section{Conclusion}
In this work, we explored the application of LLMs in concept graph recovery and question-answering within educational settings. Initially, we developed CGPrompt, enabling LLMs to perform zero-shot concept graph recovery from a given set of concepts. Subsequently, we introduced TutorQA, an expert-verified, NLP-centric benchmark designed for question-answering using a concept graph. Lastly, we devised the CGLLM pipeline, aimed at augmenting question-answering performance by leveraging LLMs.

\clearpage
\section*{Limitations}
\textbf{Graph Construction} 
Constructing a concept graph from scratch, especially in a zero-shot context, poses a significant challenge due to the quadratic increase in the number of potential concept pairs, where the possibilities grow in proportion to the square of the number of concepts. Ideally, this process would involve predictions for each pair of nodes. However, in our approach to developing the concept graph, we employed a strategy of edge sampling to minimize the number of inferences required. This decision was driven by two key considerations: firstly, we operated under the assumption that the graph exhibits sparsity, meaning that only a limited number of edges exist. Secondly, we factored in the relatively high cost associated with API usage. By adopting this methodology, we aimed to balance the thoroughness of graph construction with practical constraints on resources and expenses.

\textbf{Evaluation for TutorQA} For Tasks 2, 3, and 4, LLMs often generate novel concepts in their responses. To address this, we evaluated answers based on semantic similarities to compute an optimized F1 score. A notable limitation for both Tasks 2 and 3 is the disregard for concept order in the provided answer paths. Addressing this concern will be a focus of our future work.

\section*{Ethical Considerations}

In our research, we have meticulously addressed ethical considerations, particularly regarding our dataset TutorQA and methods CGPrompt and CGLLM. TutorQA has been expert-verified to ensure it contains no harmful or private information about individuals, thereby upholding data integrity and privacy standards. Our methods, developed on publicly available Large Language Models optimized for text generation, adhere to established ethical norms in AI research. We recognize the potential biases in such models and are committed to ongoing monitoring to prevent any unethical content generation, thereby maintaining the highest standards of research integrity and responsibility.

\bibliography{custom}
\appendix
\onecolumn

\section{Prompt Templates}
\label{app:prompt}

\subsection{Concept Graph Recovery}

\paragraph{CoT Prompt}
\mbox{}
\begin{lstlisting}
In the context of {domain}, we have two concepts: A: {concept_1} and B: {concept_2}. Assess if understanding {concept_1} is a necessary prerequisite for understanding {concept_2}. Employ the Chain of Thought approach to detail your reasoning before giving a final answer.

# Identify the Domain and Concepts: Clearly define A and B within their domain. Understand the specific content and scope of each concept.

# Analyze the Directional Relationship: Determine if knowledge of concept A is essential before one can fully grasp concept B. This involves considering if A provides foundational knowledge or skills required for understanding B.

# Evaluate Dependency: Assess whether B is dependent on A in such a way that without understanding A, one cannot understand B.

# Draw a Conclusion: Based on your analysis, decide if understanding A is a necessary prerequisite for understanding B.

# Provide a Clear Answer: After detailed reasoning, conclude with a distinct answer: <result>YES</result> if understanding A is a prerequisite for understanding B, or <result>NO</result> if it is not.   
\end{lstlisting}

\subsection{External Data in Ablation Study}
\label{app:ext_prompt}

\paragraph{CG Prompt - Doc.} 
\texttt{
We have two \{domain\} related concepts: A: \{concept\_1\} and B: \{concept\_2\}.
Do you think that people learning \{concept\_1\} will help in understanding \{concept\_2\}? \\
Hint:
1. Answer YES or NO only.
2. This is a directional relation, which means if YES, (B,A) may be False, but (A,B) is True.
3. Your answer will be used to create a knowledge graph.\\
And here are related contents to help: \{related documents concatenation\}
}

\paragraph{CG Prompt - Con.} 

\texttt{
We have two \{domain\} related concepts: A: \{concept\_1\} and B: \{concept\_2\}.
Do you think that people learning \{concept\_1\} will help in understanding \{concept\_2\}? \\
Hint:
1. Answer YES or NO only.
2. This is a directional relation, which means if YES, (B,A) may be False, but (A,B) is True.
3. Your answer will be used to create a knowledge graph.\\
And here are related contents to help: \\
We know that \{concept\_1\} is a prerequisite of the following concepts:\{1-hop successors of concept\_1 from training data\};\\
The following concepts are the prerequisites of \{concept\_1\} : \{1-hop predecessors of concept\_1 from training data\}; \\
We know that \{concept\_2\} is a prerequisite of the following concepts:\{1-hop successors of concept\_2 from training data\};\\
The following concepts are the prerequisites of \{concept\_2\} : \{1-hop predecessors of concept\_2 from training data\}. \\
}

\paragraph{CG Prompt - Wiki.} 

\texttt{
We have two \{domain\} related concepts: A: \{concept\_1\} and B: \{concept\_2\}.
Do you think that people learning \{concept\_1\} will help in understanding \{concept\_2\}? \\
Hint:
1. Answer YES or NO only.
2. This is a directional relation, which means if YES, (B,A) may be False, but (A,B) is True.
3. Your answer will be used to create a knowledge graph.\\
And here are related contents to help: \\
{Wikipedia introductory paragraph of \{concept\_1\}}\\
{Wikipedia introductory paragraph of \{concept\_2\}}
}

\subsection{CGLLM Prompt}
\textbf{Cypher Command Generation for Task1:} \\
\texttt{In the domain of natural language processing, I have already learned about "\{concept\_1\}", based on this, does it help for me to learn about "\{concept\_2\}"?\\ 
Return the path.}\\
\textbf{CGLLM Prompt:} \\
\texttt{There is a concept graph that includes the relations between concepts. \\ 
Based on the question, the path between concepts has been returned. \\
If the path is empty, then there is no relationship. \\
Only use the returned path as the information for answering. \\
Only return "Yes" or "No". \\
***Question**:\\
In the domain of natural language processing, I have already learned about \{concept\_1\}, based on this, does it help for me to learn about \{concept\_2\}?\\
***Path**:\\
\{path\}}

\section{Additional Corpora}
\label{app:tutorialbank}
\textbf{TutorialBank} We obtained the most recent version of TutorialBank from the authors, which consists of 15,583 manually curated resources. This collection includes papers, blog posts, textbook chapters, and other online resources. Each resource is accompanied by metadata and a publicly accessible URL. We downloaded the resources from these URLs and performed free text extraction. Given the varied data formats such as PDF, PPTX, and HTML, we encountered some challenges during text extraction. To ensure text quality, we filtered out sentences shorter than 25 words. Ultimately, this process yielded 559,217 sentences suitable for RAG and finetuning experiments.

\textbf{NLP Papers} We downloaded conference papers from EMNLP, ACL, and NAACL spanning the years 2021 to 2023. Following this, we utilized Grobid (\url{https://github.com/kermitt2/grobid}) for text extraction, resulting in a collection of 4,787 documents with clean text.



\section{Experimental Setup}

In our experimental setup, we employed Hugging Face's \texttt{llama-2-70b-chat-hf} (\url{https://huggingface.co/meta-llama}) model for LLaMa on a cluster equipped with 4 NVIDIA A100 GPUs. For GPT3.5 and GPT4, we used OpenAI's \texttt{gpt-3.5-turbo} and \texttt{gpt-4-1106-preview} APIs, respectively, each configured with a temperature setting of zero.
In the GCN component, utilizing LLaMa2's 8192-dimensional hidden representation, we deploy a fully connected layer to compress the input to 256 dimensions. This is then followed by GCN layers that further condense it to 128 dimensions, enabling a detailed analysis of relations and facilitating efficient learning of graph node relationships. The RAG models are implemented using Embedchain~\cite{embedchain}.

\section{More Ablation Studies}

\subsection{CG Recovery Analysis}
\label{app:cgrecovery}
In Sec.~\ref{sec:cge}, we recovered the NLP concept graph using the best zero-shot model. Theoretically, to recover a concept graph, each single pair needs to be predicted. However, given a limited budget, we conducted a sampling-based method to reduce the number of predictions. 
We initiated our investigation by sampling all existing edges in the concept graph as positive instances. An equal number of non-existent edges were also randomly selected to serve as negative instances. Utilizing zero-shot prompts on this sampled dataset, we measured link prediction performance through metrics such as Accuracy and F1 Score. The results demonstrate that GPT4 significantly excels in reinstating the Concept Graph. Specifically, in the domain of Natural Language Processing (NLP), GPT4 achieved an accuracy of 76.65\% and an F1 Score of 80.10\%. In the realm of Computer Vision (CV), it registered an accuracy of 73.22\% and an F1 Score of 76.54\%. These figures, although derived from a non-rigorous comparison, clearly demonstrate that GPT4's performance in graph recovery has reached an outstanding threshold.

\subsection{Graph Encoders}

For the supervised setting, we compare the graph convolutional network (GCN)~\cite{gatpaper} and the graph attention network (GAT) as the graph encoder. Results are shown in Tab.~\ref{app:graph_enc}. We report the best results in Tab.~\ref{tab:main}. 

\begin{table*}[h]
\centering
\begin{tabular}{@{}lcccccccc@{}}
\toprule
\textbf{Encoder} & \multicolumn{2}{c}{\textbf{NLP}} & \multicolumn{2}{c}{\textbf{CV}} & \multicolumn{2}{c}{\textbf{BIO}} & \multicolumn{2}{c}{\textbf{Overall}} \\
\cmidrule(lr){2-3} \cmidrule(lr){4-5} \cmidrule(lr){6-7} \cmidrule(lr){8-9}
 & \textbf{Acc} & \textbf{F1} & \textbf{Acc} & \textbf{F1} & \textbf{Acc} & \textbf{F1} & \textbf{Acc} & \textbf{F1} \\
\midrule
GCN     & 0.7174    & 0.7673      & 0.7839   & 0.8223     & 0.8000   & 0.8321     & 0.7671   & 0.8072 \\
GAT     & 0.5445    & 0.6498      & 0.7713   & 0.8128     & 0.8217   & 0.8464     & 0.7125   & 0.7697 \\
\bottomrule
\end{tabular}
\caption{Comparison of graph encoders.}
\label{app:graph_enc}
\end{table*}

\subsection{Finetuning}
\label{app:ft}

We further explore the impact of finetuning on additional datasets, with results detailed in Table~\ref{tab:finetuning}. Specifically, we utilize LLaMa2 70b~\cite{touvron2023llama}, finetuning it on two previously mentioned datasets: TutorialBank and NLP Papers. Both the zero-shot LLaMa and the finetuned models are employed to generate answers. As these answers are binary (\texttt{YES} or \texttt{NO}), we can calculate both the accuracy and F1 score for evaluation. However, the results indicate that finetuning does not yield positive outcomes. This can be attributed to two potential factors: 1) the poor quality of data, and 2) limited effectiveness in aiding the graph recovery task. We leave this part as the future work.

\begin{table*}[h]
\centering
\begin{tabular}{lcc}
\toprule
\textbf{Dataset} & \textbf{Acc} & \textbf{F1} \\
\midrule
LLaMa2-70b & \textbf{0.6058} & \textbf{0.6937} \\	
TutorialBank & 0.4739  & 0.0764 \\
NLP Papers & 0.5435 & 0.6363 \\
\bottomrule
\end{tabular}
\caption{Comparison of the effect of finetuning: results on NLP domain.}
\label{tab:finetuning}
\end{table*}

\section{TutorQA}
\label{app:tutorqa_info}

\subsection{Benchmark Details}
Given that TutorQA primarily focuses on the NLP domain, we utilized the same list of 322 NLP concepts and prerequisite annotations from LectureBankCD. We treated it as a directed graph and sampled from it to generate questions for Tasks 1, 2, and 3. Subsequently, domain experts reviewed these samples to filter out unsuitable questions, such as repeated cases or those involving short paths. For Task 4, we employed GPT4 to generate initial proposal examples, which were then refined and expanded upon by human experts, who also manually wrote the answers for the concept list. Finally, for Task 5, human experts carefully selected query concepts. These concepts for each question should not originate from a single path but need to be interconnected, facilitating their use in a project proposal. We show the data analysis in Tab.~\ref{tab:data_stats}.

\begin{table}[t]
\centering
\begin{tabular}{ccccccc}
\toprule
\textbf{Task} & \multicolumn{3}{c}{\textbf{Question Token}} & \multicolumn{3}{c}{\textbf{Concept/Answer Count}} \\ 
\cmidrule(lr){2-4} \cmidrule(lr){5-7} 
 & \textbf{Max} & \textbf{Min} & \textbf{Mean} & \textbf{Max} & \textbf{Min} & \textbf{Mean} \\ \midrule
T1 & 30 & 25 & 26.62 & - & - & - \\ 
T2 & 24 & 19 & 20.46 & 18 & 1 & 5.41 \\
T3 & 31 & 25 & 26.76 & 6 & 2 & 2.75 \\ 
T4 & 129 & 13 & 36.96 & 10 & 3 & 6.14 \\ 
T5 & 49 & 37 & 43.61 & - & - & - \\ \bottomrule
\end{tabular}
\caption{TutorQA Data Statistics Comparison: The answers in T1 are only "YES" or "NO", while the answers in T5 are free text with open-ended answers. }
\label{tab:data_stats}
\end{table}

\subsection{Task 4: Case Study}
We present a random example in Table~\ref{app:task4_case} to compare the performance of GPT4 and the GPT4-based CGLLM in Task 4. It is evident that GPT4 yields only one exact match along with some expanded or related concepts, whereas our CGLLM successfully generates a greater number of exact matches and equally relevant expanded concepts.

\begin{table*}[th]
\centering
\small
\begin{tabularx}{\textwidth}{lXX}
\toprule
\textbf{\textit{Question}} & \textit{In the domain of Natural Language Processing, I will give you a project title and description, tell me possible concepts that I need to learn to achieve the project. Title: Automated Legal Briefing Tool Description: A tool that automatically generates legal briefs from case files and legal documents, using NLP to extract relevant facts and arguments, streamlining the legal documentation process.} \\
\textbf{Reference} & \textcolor{bar_color1}{Event Detection; Text Summarization}; \textcolor{bar_color1}{Information Extraction}; \textcolor{bar_color1}{Natural Language Processing Introduction} \\
\midrule
\textbf{GPT4} &  \textcolor{bar_color3}{Named Entity Recognition}; \textcolor{bar_color1}{Text Summarization}, Document Classification, Sentiment Analysis, Relationship Extraction\\ 
\textbf{CGLLM} & \textcolor{bar_color3}{Named Entity Recognition}; \textcolor{bar_color1}{Information Extraction}; \textcolor{bar_color1}{Text Summarization}; Document Representation; Machine Learning Resources; \textcolor{bar_color1}{Natural Language Processing Introduction}; Neural Language Modeling; Neural Networks; \textcolor{bar_color3}{Semantic Role Labeling}; \textcolor{bar_color3}{Text Generation}; Tokenization\\
\bottomrule
\end{tabularx}
\caption{Case study on TutorQA Task 4: results on GPT4 and GPT4-based CGLLM answer. We highlight the \textcolor{bar_color1}{relevant concepts} and \textcolor{bar_color3}{expanded relevant concepts}.}
\label{app:task4_case}
\end{table*}

\subsection{Task 5: Human Evaluation Rubrics} 
\label{app:case}

\begin{itemize}
    \item \textbf{Concept Relevancy (1-5):}
    \begin{itemize}
        \item Rate how well the project description aligns with the provided query concepts.
        \item Score 1 if the project is not related to any of the query concepts.
        \item Score 5 if the project directly addresses and is entirely based on the provided query concepts.
    \end{itemize}

    \item \textbf{Concept Coverage (1-5):}
    \begin{itemize}
        \item Evaluate how comprehensively the project description covers the query concepts.
        \item Score 1 if only a minimal or superficial coverage of the concepts is present.
        \item Score 5 if all query concepts are thoroughly and extensively discussed in the project description.
    \end{itemize}

    \item \textbf{Project Convincity (1-5):}
    \begin{itemize}
        \item Assess the overall persuasiveness and viability of the project described.
        \item Score 1 if the project seems implausible, lacks a clear purpose, or doesn't appear convincing.
        \item Score 5 if the project is highly convincing, well-structured, and appears to have long-term durability.
    \end{itemize}

    \item \textbf{Scientific Factuality (1-5):}
    \begin{itemize}
        \item Determine the accuracy and correctness of the scientific details presented in the project description.
        \item Score 1 if the content contains multiple factual inaccuracies or appears entirely unscientific.
        \item Score 5 if the project description is factually accurate, well-researched, and scientifically sound.
    \end{itemize}
\end{itemize}

\subsection{Task 5 Mentioned Concept Counts} 
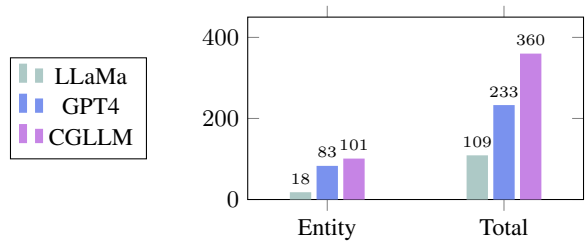
\begin{figure}
\centering
\begin{tikzpicture}
\centering
\small
\begin{axis}[
        ybar,
        ymin=0,
        ymax=450,
        bar width=8pt,
        enlarge x limits=0.45,
        legend style={
            at={(-0.3,0.5)}, 
            anchor=east,      
            legend columns=1, 
            font=\small
        },
        symbolic x coords={Entity, Total},
        xtick=data,
        nodes near coords,
        nodes near coords align={vertical},
        every node near coord/.append style={
            font=\tiny 
        },
    height=4cm,
    width=6cm,
    ]
\addplot[fill=color2, draw=none,draw opacity=0] coordinates {(Entity,18) (Total,109)};
\addlegendentry{LLaMa}

\addplot[fill=bar_color2, draw=none,draw opacity=0] coordinates {(Entity,83) (Total,233) };
\addlegendentry{GPT4}

\addplot[fill=bar_color3, draw=none,draw opacity=0] coordinates {(Entity,101) (Total,360) };
\addlegendentry{CGLLM}
\end{axis}
\end{tikzpicture}
\caption{Comparison of concept mentions in generated project proposal: LLaMa, GPT4, and CGLLM (based on GPT4). }
\label{fig:entity_comp}
\end{figure}

We also examine the diversity and novelty of the proposals predicted by the three models. One approach is to analyze the mentions of concepts. To do this, we use spaCy\footnote{https://spacy.io/} for Named Entity Recognition (NER) on all generated project proposals, and the results are compared in Fig.\ref{fig:entity_comp}. Since NER might identify many entities that are not relevant to our domain, we only retain the recognized entities that are within our list of concepts. We compare the \lq\lq Entity\rq \rq count (unique mentions of concepts) and the \lq\lq Total\rq\rq count (total mentions of concepts). We find that LLaMa is less creative; among all its proposals, it mentions only 18 unique concepts, yet with a high total count, suggesting a tendency towards suggesting homogeneous projects, like \textit{sentiment analysis}. In contrast, GPT4 and our model exhibit better performance with more diverse and domain-specific concepts, with our model proving to be the most effective. We further list the top 15 concepts that are mentioned in the three models in Tab.~\ref{tab:topic_comparison}. 



\begin{table*}[h]
\centering
\begin{tabular}{@{}lll@{}}
\toprule
\textbf{LLaMa     }                  &\textbf{ GPT4   }                        & \textbf{CGLLM}                            \\ \midrule
sentiment analysis          & sentiment analysis             & social media analysis           \\
topic modeling              & social media analysis          & sentiment analysis              \\
machine translation         & citation networks              & language modeling               \\
recurrent neural networks   & graph convolutional networks   & neural language modeling        \\
convolutional neural networks & social network extraction    & citation networks               \\
tokenization                & machine translation            & information retrieval           \\
support vector machines     & propositional logic            & text mining                     \\
syntax                      & information extraction         & neural parsing                  \\
computer vision             & semantic similarity            & robotics                        \\
preprocessing               & information retrieval          & graph convolutional networks    \\
text summarization          & document ranking               & semantic similarity             \\
question answering          & text mining                    & dimensionality reduction        \\
transfer learning           & relation extraction            & structured learning             \\
graph convolutional networks & neural language modeling     & neural networks                 \\
speech recognition          & neural machine translation     & semantic parsing                \\ \bottomrule
\end{tabular}
\caption{Comparison of concepts mentioned by different models.}
\label{tab:topic_comparison}
\end{table*}

\subsection{Task 5: More case studies}
In Table~\ref{tab:app_case_study}, we present two additional examples. Similarly to previous observations, LLaMa tends to provide much irrelevant information, whereas GPT4 and our model offer more viable project proposals. Furthermore, our model demonstrates the ability to suggest detailed concepts and incorporate related concepts as extensions. 

\begin{table*}[ht]
\centering
\small
\begin{tabularx}{\textwidth}{lXX}
\toprule
\textbf{\textit{Question 1}} & \textit{I already know about \textcolor{bar_color1}{neural question answering, recurrent neural networks and gated recurrent units}. In the domain of natural language processing, what potential project can I work on? Give me a possible idea. Show me the title and project description.} \\
\textbf{LLaMa} & \textbf{Title:} Exploring the Use of Multimodal Fusion for Enhancing Question Answering Systems; \textbf{Description:} \textcolor{bar_color1}{Question answering} systems have been a crucial component of natural language processing research in recent years. While most question answering systems rely solely on textual information, there is a growing interest in incorporating multimodal information (such as images, videos, and audio) to improve the accuracy and robustness of these systems. In this project, we aim to explore the use of \textcolor{bar_color6}{multimodal fusion} for enhancing question answering systems.  The project will involve the following tasks:  1. Collecting and annotating a large dataset of multimodal question answering examples, where each example consists of a question, a set of candidate answers, and relevant multimodal information (such as images, videos, or audio)...[\texttt{Stops because of the length restriction}]\\
\textbf{GPT4} & \textbf{Title: }Sentiment Analysis and Trend Prediction in Social Media Data;
\textbf{Description:} Leverage the power of natural language processing to analyze and predict trends in social media sentiment. Your project will focus on collecting and processing large volumes of social media posts from various platforms. Using \textcolor{bar_color3}{advanced sentiment analysis techniques}, you will classify these posts into categories like positive, negative, and neutral. The challenge is to accurately interpret not just the explicit text, but also the implied and contextual sentiment. Furthermore, \textcolor{bar_color3}{apply time-series analysis to predict future sentiment trends}, which can be invaluable for businesses, political campaigns, and public relations. This project combines NLP with data science to provide actionable insights from the vast and dynamic world of social media. \\
\textbf{CGLLM} & \textbf{Title: }Adaptive Learning Assistant for Personalized Education; \textbf{Description: }The project aims to develop an Adaptive Learning Assistant (ALA) that leverages \textcolor{bar_color1}{neural question answering, recurrent neural networks (RNNs), and gated recurrent units (GRUs)} to provide personalized educational support. By integrating \textcolor{bar_color3}{text mining techniques}, the ALA will analyze educational content and student interactions to identify knowledge gaps and learning preferences. Using document ranking, it will curate and recommend personalized learning materials and quizzes. The system will employ \textcolor{bar_color3}{character-level language models} to understand and generate nuanced responses to student inquiries, facilitating a more interactive and engaging learning experience. Additionally, incorporating a \textcolor{bar_color3}{neural Turing machine} architecture will enable the ALA to remember and utilize past interactions to improve its recommendations over time, adapting to the evolving needs of each learner. This project not only aims to enhance the educational experience by providing tailored support but also seeks to explore the boundaries of current AI capabilities in understanding and responding to complex human learning behaviors.\\
\midrule
\textbf{\textit{Question 2}} & \textit{I already know about \textcolor{bar_color1}{robotics, reinforcement learning, and agent-based view of ai}. In the domain of natural language processing, what potential project can I work on? Give me a possible idea. Show me the title and project description.} \\
\textbf{LLaMa} & \textbf{Title:} \textcolor{bar_color6}{Emotion Detection and Sentiment Analysis for Social Media Monitoring}; \textbf{Description: } In today's digital age, social media platforms have become a hub for people to express their opinions, emotions, and experiences. This project aims to develop a natural language processing (NLP) system that can analyze social media posts to detect emotions and sentiment in real-time. The system will use machine learning algorithms to classify text data into different emotional categories (e.g., happy, sad, angry, neutral) and measure the overall sentiment of each post.  The project will involve several tasks, including \textcolor{bar_color6}{data collection, data preprocessing, feature extraction, model training, and evaluation}. You will work with a large dataset of social media posts and develop a deep learning model that can accurately classify emotions and sentiment...[\texttt{Stops because of the length restriction}] \\
\textbf{GPT4} & \textbf{Title: }Autonomous Negotiation Agent for E-commerce Platforms;
\textbf{Description:}
Develop an AI-driven autonomous agent that can conduct negotiations on e-commerce platforms. This agent will use \textcolor{bar_color1}{reinforcement learning} to optimize negotiation strategies in real-time, aiming to achieve the best possible outcomes for users, whether they are buyers or sellers. The project will integrate \textcolor{bar_color6}{robotics} principles for decision-making under uncertainty and an \textcolor{bar_color6}{agent-based approach} to interact with various stakeholders. The system will analyze text data from negotiation chats, learning from past interactions to improve its bargaining skills. This tool could revolutionize e-commerce by automating and optimizing the negotiation process, potentially leading to more efficient and satisfactory transactions for all parties involved. \\
\textbf{CGLLM} & \textbf{Title: }Autonomous Robotic Search and Rescue in Hazardous Environments;  \textbf{Description: }This project aims to develop an autonomous robotic system capable of performing search and rescue missions in hazardous environments, such as after natural disasters or in industrial accidents. Utilizing the principles of \textcolor{bar_color2}{robotics}, the project will focus on creating a robot with advanced locomotion capabilities, enabling it to navigate through challenging terrains and obstacles. \textcolor{bar_color2}{The agent-based view of AI} will be central to designing the decision-making processes of the robot, allowing it to operate independently, make strategic decisions, and adapt to new information or changes in the environment. \textcolor{bar_color2}{Reinforcement learning} will be employed to train the robot to improve its search strategies over time, learning from each mission to optimize its pathfinding and victim identification processes. The project will integrate these concepts to create a system that can autonomously navigate, search, and provide immediate assistance or gather critical information in areas that are too dangerous for human rescuers, potentially saving lives and reducing the risks associated with hazardous environment interventions. \\
\bottomrule
\end{tabularx}
\caption{Two more Case studies on TutorQA Task 5: we compare the results of LLaMa, GPT4, and CGLLM. We highlight the \textcolor{bar_color1}{relevant concepts}, \textcolor{bar_color6}{irrelevant concepts},  \textcolor{bar_color2}{concept details} and \textcolor{bar_color3}{expanded relevant concepts}.}
\label{tab:app_case_study}
\end{table*}

\begin{figure*}[h]
    \centering
    \begin{tikzpicture}
    \begin{axis}[
    xbar stacked,   
    bar width=8pt,
    height=4cm,
    width = 7cm,
    enlarge y limits=0.25,
    legend style={at={(0.2,-0.05)},
      anchor=north,legend columns=-1,draw=none},
    axis y line*=none,    
    hide x axis,
    axis x line*=none,
    xlabel={Count},
    symbolic y coords={LLaMa, GPT3.5, GPT4, sup-LLaMa, sup-GPT3.5},
    ytick=data,
    yticklabel style={align=right},  
]

\addplot+[xbar,fill=color4, draw=none,draw opacity=0] plot coordinates {(52,LLaMa) (77,GPT3.5) (120,GPT4)(216,sup-LLaMa)(270,sup-GPT3.5)};
\addplot+[xbar,fill=color5, draw=none,draw opacity=0] plot coordinates {(258,LLaMa) (233,GPT3.5) (190,GPT4)(94,sup-LLaMa)(40,sup-GPT3.5)};
\legend{NEG,POS}
\end{axis}
    \end{tikzpicture}
    \caption{Positive and Negative predictions among two supervised models and three zero-shot models: results on NLP domain.}
    \label{app:neg_pos}
\end{figure*}
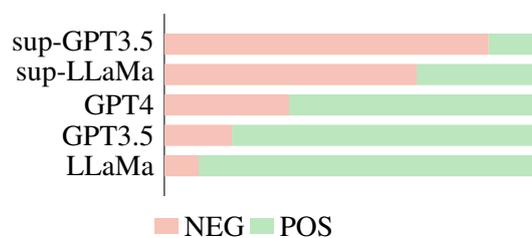

\section{Negative and Positive Predictions}

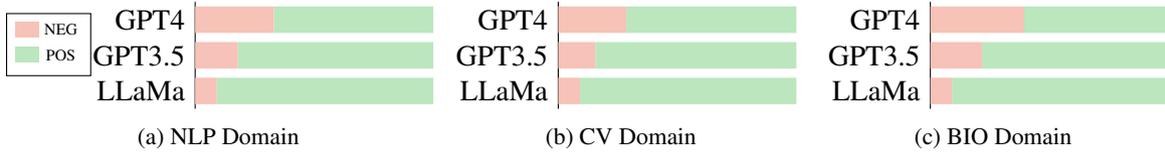
\begin{figure*}[t]
\centering
\begin{subfigure}[b]{0.35\textwidth}
    \centering
    \begin{tikzpicture}
    \begin{axis}[
    xbar stacked,   
    bar width=10pt,
    height=3cm,
    width = 5cm,
    enlarge y limits=0.25,
    legend style={
            font = \tiny,
            at={(-0.4,0.6)},   
            anchor=east,        
            legend columns=1,   
        },
    axis y line*=none,  
    hide x axis,
    axis x line*=none,
    xlabel={Count},
    symbolic y coords={LLaMa, GPT3.5, GPT4},
    ytick=data,
    yticklabel style={align=right},  
]

\addplot+[xbar,fill=color4, draw=none,draw opacity=0] plot coordinates {(52,LLaMa) (77,GPT3.5) (120,GPT4)};
\addplot+[xbar,fill=color5, draw=none,draw opacity=0] plot coordinates {(258,LLaMa) (233,GPT3.5) (190,GPT4)};
\legend{NEG,POS}
\end{axis}
    \end{tikzpicture}
    \caption{NLP Domain}
\end{subfigure}
%
\begin{subfigure}[b]{0.3\textwidth}
    \centering
    \begin{tikzpicture}
    \begin{axis}[
    xbar stacked,   
    bar width=10pt,
    height=3cm,
    width = 5cm,
    enlarge y limits=0.25,
    axis y line*=none,    
    hide x axis,
    axis x line*=none,
    xlabel={Count},
    symbolic y coords={LLaMa, GPT3.5, GPT4},
    ytick=data,
    yticklabel style={align=right},  
]
\addplot+[xbar,fill=color4, draw=none,draw opacity=0] plot coordinates {(34,LLaMa) (44,GPT3.5) (64,GPT4)};
\addplot+[xbar,fill=color5, draw=none,draw opacity=0] plot coordinates {(140,LLaMa) (130,GPT3.5) (110,GPT4)};
\end{axis}
    \end{tikzpicture}
    \caption{CV Domain}
\end{subfigure}
%
\begin{subfigure}[b]{0.3\textwidth}
    \centering
    \begin{tikzpicture}
    \begin{axis}[
    xbar stacked,   
    bar width=10pt,
    height=3cm,
    width = 5cm,
    enlarge y limits=0.25,
    legend style={at={(0.2,-0.05)},
      anchor=north,legend columns=-1,draw=none},
    axis y line*=none,    
    hide x axis,
    axis x line*=none,
    xlabel={Count},
     symbolic y coords={LLaMa, GPT3.5, GPT4},
    ytick=data,
    yticklabel style={align=right},  
]
\addplot+[xbar,fill=color4, draw=none,draw opacity=0] plot coordinates {(10,LLaMa) (15,GPT3.5) (22,GPT4)};
\addplot+[xbar,fill=color5, draw=none,draw opacity=0] plot coordinates {(36,LLaMa) (31,GPT3.5) (24,GPT4)};

\end{axis}
    \end{tikzpicture}
    \caption{BIO Domain}
\end{subfigure}

\caption{Comparison of Positive and Negative predictions on all domains.}
\label{fig:pos_neg}
\end{figure*}

To gain a deeper understanding and comparison of model performance, we complement the results from Table~\ref{tab:main} with an analysis of the confusion matrix, summarizing the counts of Negative and Positive predictions in Figure~\ref{fig:pos_neg}. This analysis is based on a random subset of the 5-fold validation set, ensuring a fair comparison. Across all domains and models, we observe a pronounced tendency towards Positive predictions over Negative ones. This suggests a bias in the models towards affirming the existence of a relationship in given concept pairs. One possible explanation is that all concepts belong to a single domain, possibly leading LLMs to infer potential relationships inherent to the domain. Despite providing a clear definition of \lq relationship', even the most adept model, GPT4, exhibits a high rate of Positive predictions. Moreover, in Fig.~\ref{app:neg_pos}, focusing on the NLP domain, we present results from a supervised setting: \textbf{sup-GPT3.5} (a supervised graph model with GPT3.5 embeddings) and \textbf{sup-LLaMa} (a supervised graph model with LLaMa embeddings). The performance of sup-GPT3.5 is notably poor, with an accuracy of 0.5368 and an F1 score of 0.6639, and it shows a tendency to predict more negative labels.

\end{document}